\documentclass[conference]{IEEEtran}
\IEEEoverridecommandlockouts
\usepackage{cite}
\usepackage{amsmath,amssymb,amsfonts,array,multirow}
\usepackage{algorithm} 
\usepackage{algorithmic}
\usepackage{graphicx}
\usepackage{epstopdf}
\usepackage{textcomp}
\usepackage{xcolor}

\usepackage{booktabs}

\def\BibTeX{{\rm B\kern-.05em{\sc i\kern-.025em b}\kern-.08em
    T\kern-.1667em\lower.7ex\hbox{E}\kern-.125emX}}

\begin{document}


\title{Cross-subject Action Unit Detection with Meta Learning and Transformer-based Relation Modeling}

\author{\IEEEauthorblockN{Jiyuan Cao$^{1}$, Zhilei Liu$^{1*}$\thanks{*Corresponding author}, Yong Zhang$^{2}$}
\IEEEauthorblockA{$^{1}$ College of Intelligence and Computing, Tianjin University, Tianjin, China\\
$^{2}$ Tencent AI Lab, Shenzhen, China\\
\{caojiyuan, zhileiliu\}@tju.edu.cn, zhangyong201303@gmail.com
}
}

\maketitle

\begin{abstract}
Facial Action Unit (AU) detection is a crucial task for emotion analysis from facial movements. The apparent differences of different subjects sometimes mislead changes brought by AUs, resulting in inaccurate results. However, most of the existing AU detection methods based on deep learning didn't consider the identity information of different subjects. The paper proposes a meta-learning-based cross-subject AU detection model to eliminate the identity-caused differences. Besides, a transformer-based relation learning module is introduced to learn the latent relations of multiple AUs. To be specific, our proposed work is composed of two sub-tasks. The first sub-task is meta-learning-based AU local region representation learning, called MARL, which learns discriminative representation of local AU regions that incorporates the shared information of multiple subjects and eliminates identity-caused differences. The second sub-task uses the local region representation of AU of the first sub-task as input, then adds relationship learning based on the transformer encoder architecture to capture AU relationships. The entire training process is cascaded. Ablation study and visualization show that our MARL can eliminate identity-caused differences, thus obtaining a robust and generalized AU discriminative embedding representation. Our results prove that on the two public datasets BP4D and DISFA, our method is superior to the state-of-the-art technology, and the F1 score is improved by 1.3\% and 1.4\%, respectively.
\end{abstract}


\section{Introduction}
\par Facial action unit system (FACS) \cite{ekman1997face} is a comprehensive and objective system to describe facial expressions. It defines a unique set of basic facial muscle movements, called action units (AU). Through the combination of AUs, any possible facial expressions can be described, and emotions and cognitive states can be further explained \cite{martinez2017automatic}. In recent years, facial action unit (FAU), as a comprehensive description of facial movements, has been paid more and more attention in the fields of human-computer interaction and emotional computing. The task of FAU detection can be expressed as a multi-label binary classification problem of detecting each AU \cite{li2019semantic, zhao2016deep}. Facial AU detection is beneficial for the recognition and analysis of facial expressions.
\par
\begin{figure}[htbp]
\centering
\includegraphics[width=0.45\textwidth]{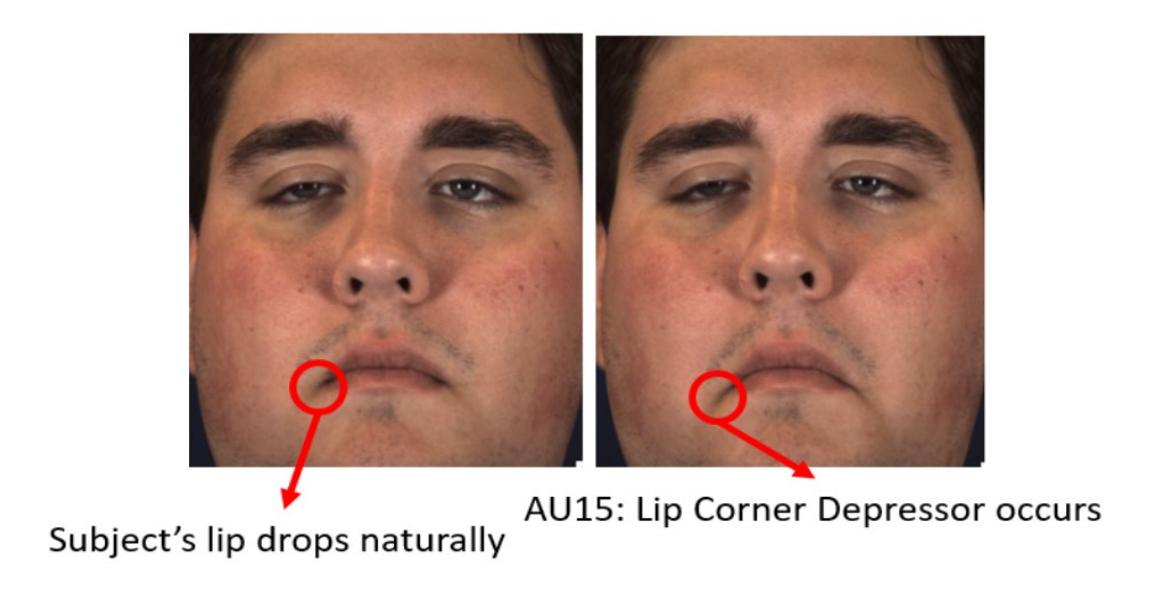}
\caption{Action units differentiated by identity}
\label{fig:idf}
\end{figure}

 Due to differences in races, genders, ages, etc, subjects' appearance differences sometimes mislead changes brought by local AU. For example, as shown in Figure \ref{fig:idf}, the neutral face of this person with naturally drooping lips looked like he had AU15 (lip corner depressor), but in fact he did not. It can be seen that there is a problem of misleading AU detection due to identity differences, but so far only a few articles have paid attention to this problem. Tu et al. \cite{tu2019idennet} proposed IdenNet to solve the problem of AU detection considering identity differences, in which the adopted CNN architecture cascades two separate tasks, one for identity-dependent feature extraction and the other for AU detection with identity subtraction. However, since the learning of facial feature space in the first sub-task depends on manually selected triplets, the method is not adaptive. In addition, the proposed model did not consider the regional characteristics of AU with local activation. 

In this paper, our first sub-task mainly addresses the problem of identity-caused differences while considering the properties of AU local activation, so we propose a method which is a meta-learning based AU local region representation learning, called \textbf{MARL}.
We notice that meta-learning has the characteristic of merging shared information of multiple tasks \cite{hospedales2020meta}. Corresponding to AU detection, if each subject is regarded as a task and the AU detection model is trained using the meta-learning framework, the shared information of multiple subjects can be integrated and identity-caused differences can be solved naturally. Moreover, since AU has the characteristics of regional activation, that is, AU occurs in the local region of the face, inspired by \cite{li2019semantic}, we propose an AU region model to pay more attention to the local AU region learning. 
After solving the problem of identity differences, we want to further optimize the model to achieve state-of-the-art, thus we consider adding relation learning. A common sense in the field of AU detection is that AUs are not independent of each other, because a facial expression activates multiple AUs simultaneously. 
In recent years, relational learning has been considered in many AU detection works \cite{li2017action, niu2019local, li2019semantic, liu2020relation, jacob2021facial}, which indicates that relational learning is one of the research points to further improve the results of AU detection in current AU detection research. Among them, Jacob et al. \cite{jacob2021facial} use the transformer for relation learning, and its detection results outperform other methods. This work verifies that the transformer can effectively capture AU relations. However, the extracted AU features still do not consider identity differences.
In the second sub-task of this paper, we consider using a transformer encoder to capture the relationship between different action units by self-attention, differs from \cite{jacob2021facial}, AU features as input has eliminated identity differences. Ablation studies also demonstrate that the transformer can capture AU relationships to significantly improve detection results.
\par To sum up, the contributions of this paper are threefold:
\begin{enumerate}
    \item We propose a method which is an AU local region representation learning based on meta-learning. It can integrate the shared information of multiple subjects and eliminate identity-caused differences. 
    \item We propose an AU relation learning module based on a transformer-encoder architecture, in which the input is AU features with identity differences removed. This module can effectively capture the AU relationship. 
    \item We conduct extensive experiments on the widely used BP4D and DISFA datasets and demonstrate the superiority of the proposed method over the state-of-the-art facial AU detection methods.
\end{enumerate}

\section{RELATED WORK}
\par Our proposed framework is mainly related to facial AU detection and meta-learning, thus we divide related methods into two groups according to the way they approach the problem, and discuss our relationship with existing methods.
\subsection{Facial Action Unit Detection}
\par Zhang et al. \cite{zhang2018identity} leveraged identity annotation datasets to enrich face variants during AU detection learning. They proposed an adversarial training framework (ATF) that confused CNNs when classifying identities to produce subject-invariant features. Tu et al. \cite{tu2019idennet} proposed IdenNet, which aims to solve the AU detection problem of identity-caused differences. They adopt the architecture of CNN cascades containing two separate tasks, for identity-dependent feature extraction in one task and identity subtraction along with AU detection in another task. However, their method requires manual selection of distinct triplets, and can not automatically learn identity differences information. Compared to the above methods, our MARL combines a meta-learning framework with good performance on learning shared information of multiple subjects to build robust features for AU detection. 
\par Li et al. \cite{li2017eac} propose EAC-Net for facial AU detection by enhancing and cropping predefined ROIs for each AU. All ROIs with center locations specified by landmarks have fixed size and fixed attention distribution. Li et al. \cite{li2019semantic} further refer to EAC-Net and employ a separate filter in each cropped AU region to train its representation separately, and finally merges the same AUs according to symmetry to obtain semantically guided AU features. In view of the success of these methods, we also use landmarks to extract regions of interest for AU region learning.
\par Several works explicitly take the relationships of AUs into account. Shao et al. \cite{shao2019facial} proposed ARL, using Conditional Random Field (CRF) to model pixel-level relations so as to refine the initial spatial attention weight. Li et al. \cite{li2019semantic} proposed SREAL, AU correlation is modeled by using a knowledge graph of AU relationships and a Gated Graph Neural Network. Similarly, Liu et al. \cite{liu2020relation} proposed AU-GCN using graph convolutional network (GCN) for AU relation modeling. Both \cite{li2019semantic} and \cite{liu2020relation} learn the spatial relationships of the AUs in face images. Inspired by Attention Branch Networks (ABN) \cite{fukui2019attention} and Transfomer \cite{vaswani2017attention}, Jacob et.al \cite{jacob2021facial} proposed AU detection with transformer, they used attention branch networks combined with feature maps to obtain multiple AU feature embeddings with attention information, and then added transformers to capture the relationship between AU feature embeddings. The state-of-the-art results demonstrate that the transformer can effectively learn AU relations. Inspired by the transformer \cite{vaswani2017attention, jacob2021facial}, this paper introduces an AU correlation module based on the transformer encoder to capture the AU relationship. 

\subsection{Meta Learning}
\par Meta-learning, or learning to learn, is the science of systematically observing how different machine learning approaches perform on a wide range of learning tasks, and then learning from this experience, or meta-data, to learn new tasks much faster than otherwise possible \cite{vanschoren2018meta}. Currently, meta-learning studies have focused on learning good weight initialization for few-shot learning \cite{finn2017model}, or learning to generate the optimal hyper-parameters and optimizer based on a meta net \cite{ravi2016optimization, andrychowicz2016learning} or learning to generate useful auxiliary labels \cite{liu2019self}. Model-Agnostic Meta-Learning (MAML) \cite{finn2017model}, a meta-learning method based on gradient descent, finds the optimal gradient descent direction by learning the gradients of multiple tasks, thereby guiding the meta-learner to learn the meta-knowledge of multiple tasks. MAML has the meta-learner who learns the differences across training tasks and the learner who learns and optimizes each of the tasks itself. Comparing other few-shot learning methods such as \cite{ravi2016optimization} which needs an additional network as LSTM to train a classifier or \cite{koch2015siamese} which obtains feature embedding with the Siamese network and adds a non-parametric method for a new task, MAML needs only one single target network to train meta-learner. Any model network can be used as the one and only backbone network, which makes it called model-agnostic. Corresponding to identity-aware AU detection, we can regard the subject as a task and the AU model as a meta-learner, and use the MAML training framework to train the model without adding additional network structures, and finally get the meta parameters, which can eliminate identity-caused differences.

\begin{figure}[htbp]
\centering
\includegraphics[width=0.3\textwidth]{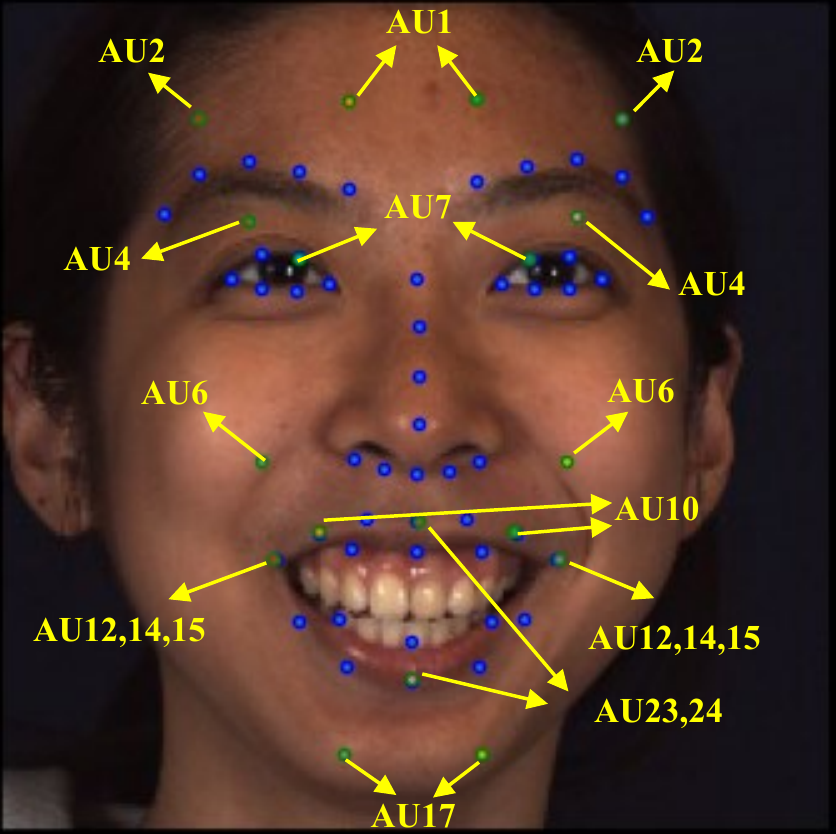}
\caption{The location of facial AU centers.}
\label{fig:landmarks}
\end{figure}

\section{Proposed Method}
\subsection{Overview}
\begin{figure*}[htbp]
\centering
\includegraphics[width=0.95\textwidth]{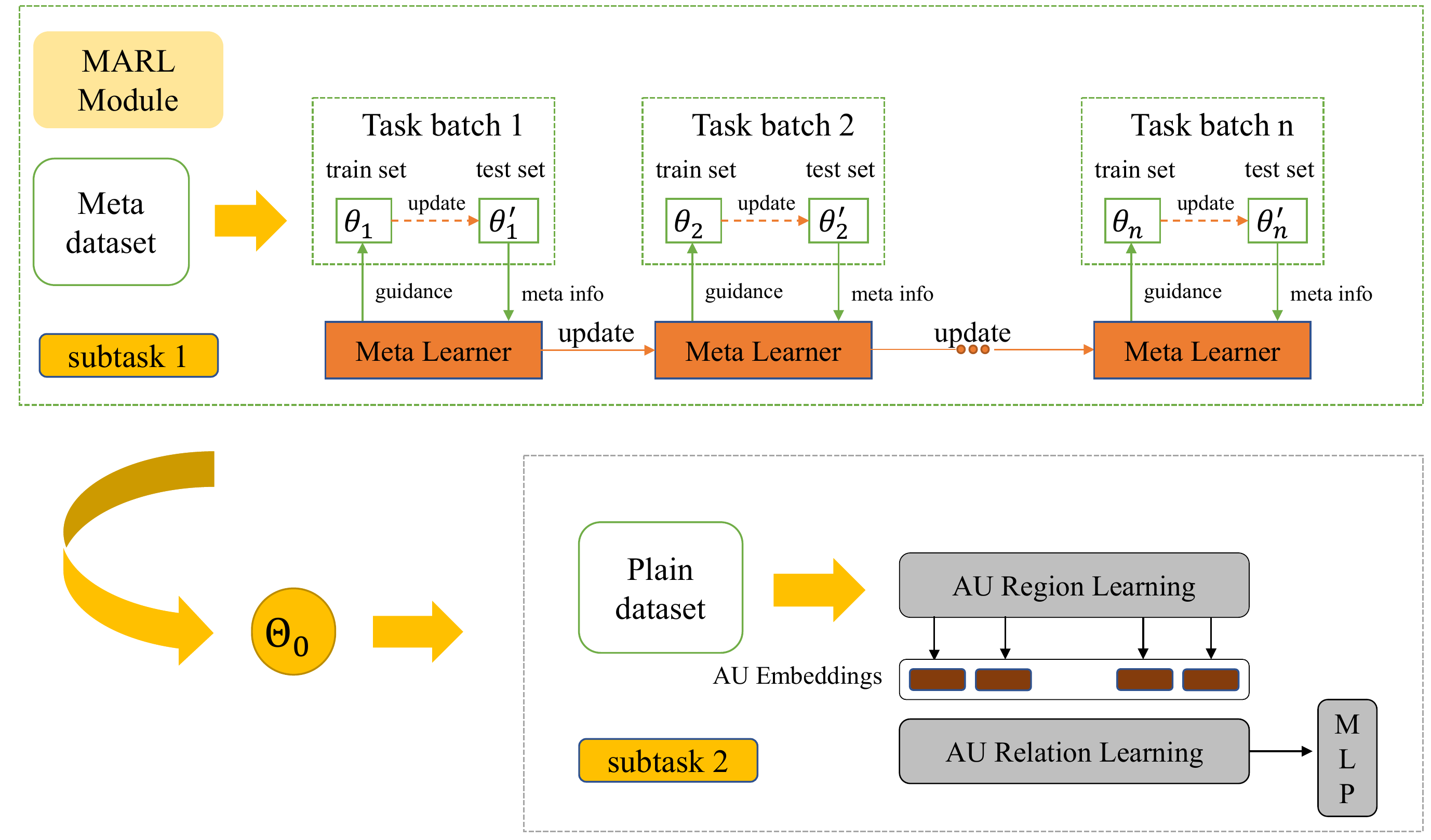}
\caption{Overall Architecture. Our proposed method is implemented in an architecture of multi-task network cascades.}
\label{fig:frameworks}
\end{figure*}

Our proposed method is implemented in an architecture of multi-task network cascades shown in Figure \ref{fig:frameworks}. Specifically, our proposed method mainly consists of two sub-tasks. The first sub-task is AU local region representation learning based on meta-learning, which uses a region learning network (shown in Figure \ref{fig:RL}) as a meta-learner to learn a meta-parameter that incorporates the shared information of multiple subjects, namely $\Theta_0$, as the pre-training parameters of the second sub-task. During the training process, the dataset is processed into a meta-dataset according to the settings of meta-learning. 
The second sub-task is relational learning based on the transformer encoder architecture. The input uses the plain dataset, we import the meta-parameters of the first sub-task to generate AU embeddings that integrate the commonalities of multiple subjects and eliminate the identity-caused differences. Then these AU embeddings are used as input, the AU relationship is captured by the transformer encoder, and finally, the multi-label prediction is executed. The details of these two components are introduced in the following.

\subsection{Meta-Learning based AU Local Region Representation Learning}
\label{sec:SecB}
\subsubsection{\textbf{AU Local Region Representation Network}}
\par
\begin{figure*}[htbp]
\centering
\includegraphics[width=0.95\textwidth]{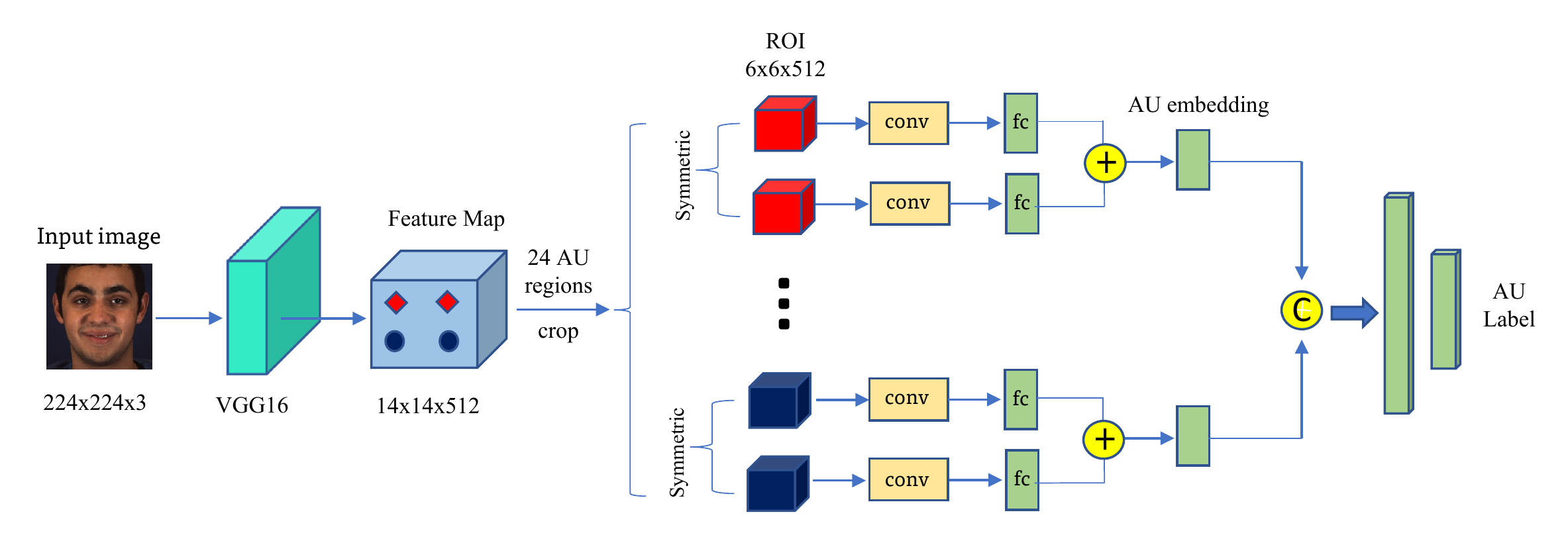}
\caption{AU Local Region Representation Network. We treat this network as a meta-learner for meta-learning.}
\label{fig:RL}
\end{figure*}

 Inspired by SRERL \cite{li2019semantic}, we adopt a similar network structure with fewer net parameters in this paper which is shown in Figure \ref{fig:RL}. We choose VGG16 \cite{simonyan2014very} as our backbone network, which is composed of 5 groups of convolutional layers with down-sampling. Due to the trade-off between feature performance and resolution, we use the first four groups of convolutional layers to extract features. The input of the feature extractor is a $224\times224$ RGB facial image with its landmark information. After this, a feature map with a size of $14\times14$ and a channel number of 512 is obtained. Specifically, we capture adaptive AU regions by using facial landmarks. Figure \ref{fig:landmarks} illustrates the correspondence between facial AUs and facial landmarks. Since there is a relationship between facial anatomical regions and AUs, we can generate the location of the center of each AU, which further corresponds to the nearest facial landmark. Then we use the AU centers located by landmarks to crop a $6\times6$ region from the concatenated global feature maps for its initial regional feature. We further refer to \cite{li2017eac} and adopt separate filters in each cropped AU region to train its representation separately. Suppose we have $C$ AUs. Due to the symmetry of the face, we obtain $2C$ patch-wise feature maps (each AU corresponds to two patches), and design $2C$ independent regional feature learning branches. For each individual region learning branch, we use a $14\times14$ convolutional layer and a fully connected layer to learn the specific local representation. These convolutional and fully connected layers are trained separately, which helps to avoid information interference between different AU regions caused by receptive fields of different scales.

\subsubsection{\textbf{Meta-Learning Algorithm for AU Region Learning}}
Our goal is to learn optimal meta-parameters that incorporates shared information from multiple subjects, therefore the training strategy of meta-learning is adopted. We use the gradient descent based meta-learning algorithm MAML \cite{finn2017model} to train the model, because it finds the optimal gradient descent direction by learning the gradients of multiple tasks, thereby guiding the fusion of shared knowledge across multiple tasks. Algorithm \ref{alg:alg1} illustrates the overall training sequence of our model based on MAML.

\begin{algorithm}
    \renewcommand{\algorithmicrequire}{\textbf{Input:}}
	\renewcommand{\algorithmicensure}{\textbf{Output:}}
    \caption{Meta-Learning Algorithm For AU Region Learning}
    \label{alg:alg1}
    \begin{algorithmic}[1]
        \REQUIRE
            Meta-Training Set $\left\{{D_s^{T_i}, D_q^{T_i}}\right\}_{i=1}^N$;
            Meta-Testing Set $\left\{{D_a, D_t}\right\}$;
            $\alpha$, $\beta$ step size hyper parameters.
        \ENSURE
            Optimal Meta-Learner parameters $\Theta_0$.
        \STATE Initialize Meta-Learner parameters $\Theta_0$ using pre-train backbone VGG-Face Net.
        \REPEAT
            \STATE $//$ Meta-Training
            \STATE Sample task batch = $\left\{{D_s^{T_i}, D_q^{T_i}}\right\}_{i=1}^K$
            \FOR{i=1,2,...,K}
                \STATE Let $\theta=\Theta_0$
                \STATE Evaluate $L_{inner}$ using $\theta$ and $D_s^{T_i}$
                \STATE Update $\theta_i=\theta-\alpha\nabla_\theta{L_{inner}}$
                \STATE Evaluate $L_{outer}$ using updated $\theta_i$ and $D_q^{T_i}$
            \ENDFOR
            \STATE $\theta\leftarrow\theta-\frac{\beta}{k}{\sum_{i=0}^k}{L_{outer}}$, $\Theta_0=\theta$
            \STATE $//$ Meta-Testing
            \STATE Sample Meta-Testing set $\left\{{D_a, D_t}\right\}$
            \STATE Let $\theta_{te} = \Theta_0$, update $\theta_{te}$ using $D_a$, get adapt model $f_{\theta_{te}}$
            \STATE Input $D_t$ into $f_{\theta_{te}}$ to get result. 
            \STATE Record best result and save optimal $\Theta_0$ each time.
        \UNTIL{Convergence}
	\end{algorithmic} 
\end{algorithm}

\par According to the setting of meta-learning, we need to build a meta-dataset first. Similar to the data batch sampling method of supervised learning, meta-learning uses the task batch sampling method, i.e., a batch of tasks is sampled for training at each time. We treat each subject as a task, and sample $B$ tasks as a batch each time. For each task, all the samples are divided into support set with $S$ samples and a query set with $Q$ samples respectively by using a random sampling strategy with $N$ batches per epoch. 
\par We treat AU local region representation network as meta-learner, and use the constructed meta-dataset as input, then we build the meta-learning training framework based on MAML. In combination with algorithm \ref{alg:alg1}, we describe the training process of meta-learning in detail, as shown below:
\begin{enumerate}
    \item[a)] \textbf{Meta-Training}
        \begin{enumerate}
            \item[1)] Initialize the parameters of the meta-learner. We denote the initialization parameter as $\Theta_0$, and use it as the initial incoming parameter in the update phase of the inner loop of meta-learning, that is, $\theta=\Theta_0$.
            \item[2)] Inner loop training update. Import task batch data, each task batch can be represented as task batch = $\left\{{D_s^{T_i}, D_q^{T_i}}\right\}_{i=1}^K$. Input the support set data $D_s$ to the meta-learner, and perform forward propagation on the current model parameters $\theta$. Calculate the loss $L_s$ between the prediction result of forward propagation and the training label, solving the gradient of the current loss $L_s$ relative to the model parameter $\theta$ and update the model parameter $\theta$ according to the obtained gradient value.
            \item[3)] Outer loop training update. The query set data $D_q$ is used as input, forward pass is performed on model parameters $\theta$ and the loss difference $L_q$ between the prediction result of forward propagation and the training label is calculated.
            The current query set loss $L_q$ taken as the outer loop loss of the meta-learner, that is $L_{outer}$, and the initial model parameter $\Theta_0$ of the meta-learner is updated.
        \end{enumerate}
    \item[b)] \textbf{Meta Testing}
        \begin{enumerate}
            \item[4)] During testing, the test set is divided into an adaptation set and a query set, namely test batch = $\left\{{D_a, D_q}\right\}$, use $D_a$ as the input, repeat step $1)$ to $2)$, the purpose is to enable the model to quickly adapt to the new model after several steps of gradient descent, and then directly use the model parameters output in step $2)$ to test $D_q$, forward propagation to get the test results, and complete the result prediction. 
        \end{enumerate}
\end{enumerate}

\par In each iteration, the model parameters $\Theta_0$ updated in step $3)$ are used as the initialization parameters of the meta-learner, the task batch sampling is performed again, and steps $1)$ to $3)$ are repeated many times until the meta-learner converges. The purpose of meta-testing (step 4) is to filter out the meta-model with the best detection results, then we save the meta-model parameters with the best test results so that they can be imported as pre-training parameters in the subsequent relation learning sub-task.

\subsection{Relation Learning based on Transformer Encoder Architecture}
\label{sec:SecC}
\begin{figure}[htbp]
\centering
\includegraphics[width=0.4\textwidth, height=0.6\textwidth]{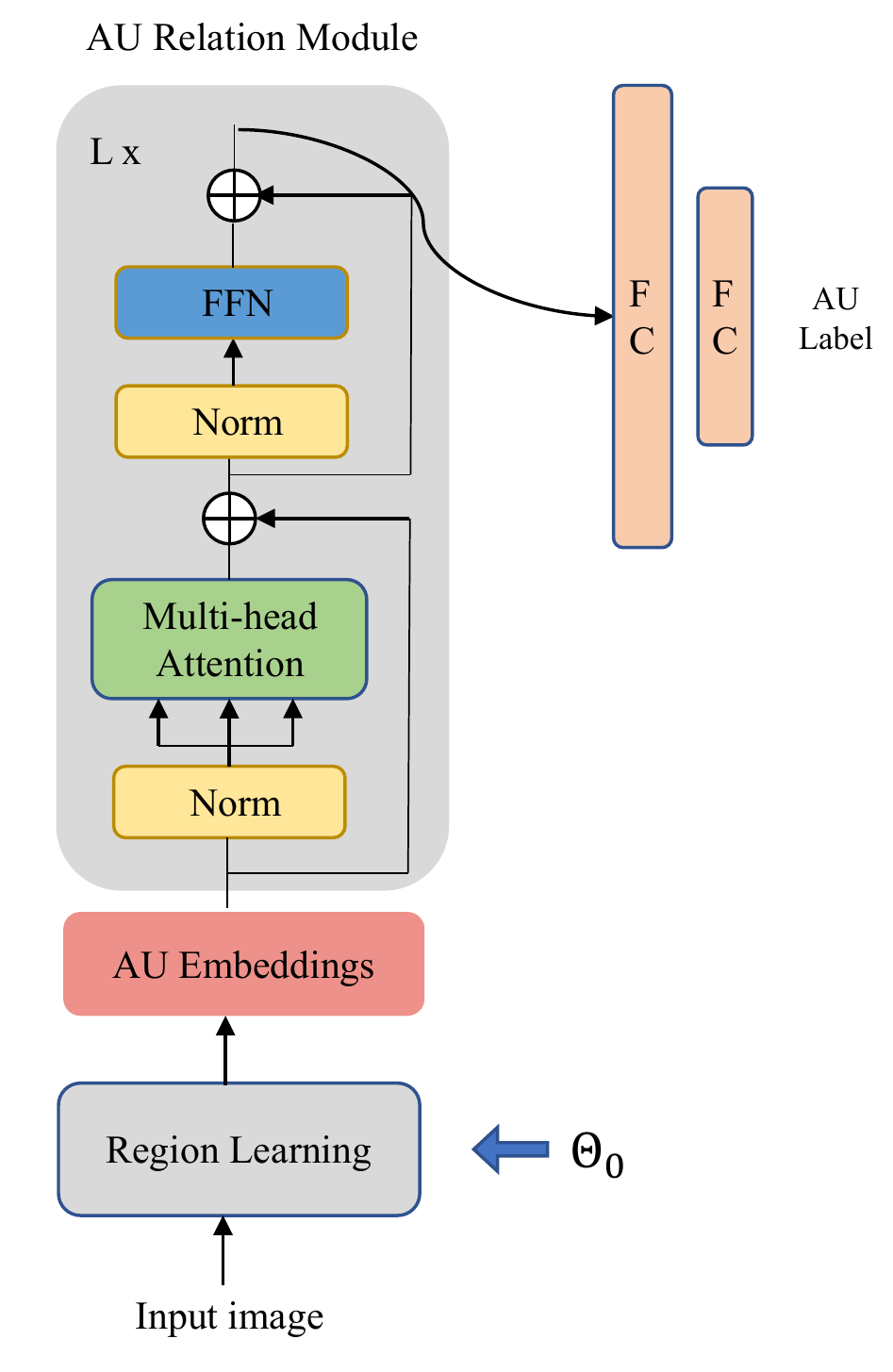}
\caption{AU Relation Learning Module with Transformer Encoder.}
\label{fig:relation}
\end{figure}

The details of this module are illustrated in Figure \ref{fig:relation}. The AU relation module estimates the relationships between the discriminative AU embeddings. First, we remove the last two fully connected layers of the region representation network, then import the optimal meta-model parameters saved from the previous meta-learning training, and use them as pre-training parameters to obtain AU embeddings. This unit consists of a transformer encoder \cite{vaswani2017attention}, which takes the AU embeddings as input. The transformer encoder has two main components: multi-head attention and a feed-forward network (FFN), with normalization layers in between.
\cite{vaswani2017attention}. The features from the AU relation learning module are passed through a classifier with two fully connected layers to obtain the predicted labels. The output from the classifier is taken as the predicted labels during the time of inference. Note that at this point we use the plain dataset instead of the meta-dataset for supervised learning.

\subsection{Loss Function}
\par Facial AU detection can be regarded as a multi-label binary classification problem with the following weighted multi-label cross entropy loss: 
\begin{equation}\label{eq:Lbce}
L_{bce} = -\sum_{i=1}^{n_{au}} w_i [p_{i} \log \hat{p}_{i} + (1-p_{i}) \log (1-\hat{p}_{i})],
\end{equation}
where $p_{i}$ denotes the ground-truth occurrence probability of the $i$-th AU, which is $1$ if occurrence and $0$ otherwise. The weight $w_i$ introduced in Eq.\eqref{eq:Lbce} is to alleviate the data imbalance problem. For most facial AU detection benchmarks, the occurrence rates of AUs are imbalanced. Since AUs are not mutually independent, imbalanced training data has a bad influence on this multi-label learning task. Particularly, we set $w_i = (1/r_i)/\sum_{u=1}^{n_{au}}(1/r_u)$, where $r_i$ is the occurrence rate of the $i$-th AU in the training set.
In many cases, some AUs appear rarely in training samples, for which the cross entropy loss in Eq.\eqref{eq:Lbce} often makes the AU prediction strongly biased towards non-occurrence. To address this, we exploit precision and recall which are both relevant to the true positive. Since F1-score: $F1=2PR/(P+R)$ considers both precision $P$ and recall $R$, we introduce a weighted multi-label Dice coefficient (F1-score) loss \cite{milletari2016v}:
\begin{equation}\label{eq:Lf1}
L_{f_1} = \sum_{i=1}^{n_{au}} w_i (1-\frac{2 p_{i} \hat{p}_{i} + \epsilon}{p_{i}^2 + \hat{p}_{i}^2 + \epsilon}),
\end{equation}
where $\epsilon$ is a smooth term. F1-score is known as the most popular evaluation metric for facial AU detection. The use of Eq.\eqref{eq:Lf1} keeps the consistency between the learning process and the evaluation metric. By combining Eq.\eqref{eq:Lbce} and Eq.\eqref{eq:Lf1}, we can obtain the overall AU detection loss:
\begin{equation}\label{eq:Lau}
L_{au} = L_{bce} + \mu{L_{f_1}},
\end{equation}
where the occurrence probability $\hat{p}_i$ of each AU is predicted based on the integrated information of all the AUs.

\par In Section B of this chapter, we define $L_{inner}$ in Eq.\eqref{eq:Lbce} and $L_{outer}$ in Eq.\eqref{eq:Lau}. In Section C, we define loss between predicted labels and ground-truth labels is Eq.\eqref{eq:Lau}.

\begin{table}[htbp]
\centering\caption{Results on the BP4D dataset. Comparison with state-of-the-art methods using the F1-score metric.}
\begin{tabular}{*{8}{c}} 
 \toprule
 AU&EAC&DSIN&JAA&LP&ARL&SREAL&Ours\\ 
 \midrule
 1&39.0&\textbf{51.7}&47.2&43.4&45.8&46.9&46.0\\
 2&35.2&40.4&44.0&38.0&39.8&\textbf{45.3}&43.2\\
 4&48.6&\textbf{56.6}&54.9&54.2&55.1&55.6&52.1\\
 6&76.1&76.1&77.5&77.1&75.7&77.1&\textbf{82.5}\\
 7&72.9&73.5&74.6&76.7&77.2&78.4&\textbf{79.9}\\
 10&81.9&79.9&84.0&83.8&82.3&83.5&\textbf{86.0}\\
 12&86.2&85.4&86.9&87.2&86.6&87.6&\textbf{90.9}\\
 14&58.8&62.7&61.9&63.3&58.8&63.9&\textbf{65.0}\\
 15&37.5&37.3&43.6&45.3&47.6&\textbf{52.2}&48.0\\
 17&59.1&62.9&60.3&60.5&62.1&\textbf{63.9}&63.1\\
 23&35.9&38.8&42.7&48.1&47.4&47.1&\textbf{50.6}\\
 24&35.8&41.6&41.9&54.2&\textbf{55.4}&53.3&54.0\\
 \midrule
 Avg.&55.9&58.9&60.0&61.0&61.1&62.1&\textbf{63.4}\\
 \bottomrule
\end{tabular}
\label{table:BP4D}
\end{table}

\begin{table}[htbp]
\centering\caption{Results on the DISFA dataset. Comparison with state-of-the-art methods using the F1-score metric.}
\begin{tabular}{*{8}{c}} 
 \toprule
 AU&EAC&DSIN&JAA&LP&ARL&SREAL&Ours\\ 
 \midrule
 1&41.5&42.4&43.7&29.9&43.9&\textbf{45.7}&34.7\\
 2&26.4&39.0&46.2&24.7&42.1&\textbf{47.8}&31.5\\
 4&66.4&68.4&56.0&72.7&63.6&59.6&\textbf{76.1}\\
 6&\textbf{50.7}&28.6&41.4&46.8&41.8&47.1&50.6\\
 9&\textbf{80.5}&46.8&44.7&49.6&40.0&45.6&53.2\\
 12&\textbf{89.3}&70.8&69.6&72.9&76.2&73.5&73.8\\
 25&88.9&90.4&88.3&93.8&95.2&84.3&\textbf{93.7}\\
 26&15.6&42.2&58.4&65.0&66.8&43.6&\textbf{67.1}\\
 \midrule
 Avg.&48.5&53.6&56.0&56.9&58.7&55.9&\textbf{60.1}\\
 \bottomrule
\end{tabular}
\label{table:DISFA}
\end{table}

\begin{table*}
\centering
\caption{Ablation Study On BP4D dataset by using the F1-score metric}
\begin{tabular}{cccccc} 
\toprule
Method&Pre-Trained&Region Learning&Meta Learning&Encoder&F1-score\\ 
\midrule
baseline&-\-&-&-&-&55.2\\
PT&\checkmark&-&-&-&56.5\\
PT-RL&\checkmark&\checkmark&-&-&60.7\\
PT-RL-E&\checkmark&\checkmark&-&\checkmark&61.8\\
\midrule
PT-ML&\checkmark&-&\checkmark&-&57.9\\
PT-RL-ML&\checkmark&\checkmark&\checkmark&-&61.7\\
\midrule
PT-RL-ML-E&\checkmark&\checkmark&\checkmark&\checkmark&63.4\\
\bottomrule
\end{tabular}
\label{table:abs}
\end{table*}

\section{EXPERIMENTS}
\subsection{Datasets and Settings}
\subsubsection{Datasets}
\par Our method is evaluated on two widely used datasets for AU detection, i.e.: BP4D \cite{zhang2014bp4d} and DISFA \cite{mavadati2013disfa}, in which both AU and landmark labels are provided.
\par \textbf{-BP4D} contains 41 subjects with 23 females and 18 males, each of which is involved in 8 sessions. There are 328 videos, including about 140,000 frames, with AU labels that occurrence or non-occurrence. Each frame is also annotated with 49 landmarks detected by SDM \cite{de2015intraface}. Similar to the settings of \cite{zhao2016deep}, \cite{li2017eac}, 12 AUs (1, 2, 4, 6, 7, 10, 12, 14, 15, 17, 23, and 24) are evaluated using subject exclusive 3-fold cross-validation, where two folds are used for training and the remaining one is used for testing.
\par \textbf{-DISFA} consists of 27 videos recorded from 12 women and 15 men, each of which has 4,845 frames. Each frame is annotated with AU intensities on a six-point ordinal scale from 0 to 5 and 66 landmarks detected by AAM \cite{cootes2001active}. To be consistent with BP4D, we use 49 landmarks, a subset of the 66 landmarks. It has a serious data imbalance problem, where most AUs have very low occurrences and only a few other AUs have high occurrences. According to the setting in \cite{zhao2016deep}, \cite{li2017eac}. (2018), AU intensities equal to or greater than 2 are considered as occurrence, while others are treated as non-occurrence. Subject exclusive 3-fold cross-validation is also conducted with evaluations on 8 AUs (1, 2, 4, 6, 9, 12, 25, and 26).
\subsection{Implementation Details}
The inputs are resized to $224\times224$ RGB images. The dataset is divided into 3 folds according to the number of subjects, 2 folds for training and 1 fold for testing, we use three-fold cross-validation. $n_{au}$ is 12 and 8 for BP4D and DISFA respectively. The trade-off parameter $\mu$=1.5 is set to balance loss, and smooth term $\epsilon$ is 1. Our project is implemented using PyTorch.
\par For the MARL module, we use the meta-learning setting to train 100 epochs, each epoch samples 100 task batches, each task batch samples the number of tasks K=5, each task samples S=5, Q=15 for a total of 20 images. In meta-testing, we sample 600 task batches, the sampling setting is the same as meta-training. The outer loop uses Adam as the optimizer and the learning rate is set to $\beta=0.006$, and the learning rate of the inner loop is $\alpha=0.01$. For the AU relation learning module, we use supervised learning to train 30 epochs, the training data batch size is 16, and the testing data batch size is 32, using Adam with an initial learning rate of 0.006, in which the learning rate is multiplied by a factor of 0.3 at every 2 epochs.
\subsection{Evaluation Metrics}
Following the previous methods of \cite{zhao2016deep, li2017eac}, the frame-based F1-score (F1-frame, \%) is reported. In addition, we compute the average results overall AUs (Avg). In the following sections, we omit \% in all the results for simplicity.
\subsection{Comparison with State-of-the-Art Methods}
We compare our method against state-of-the-art single-frame-based facial AU detection methods under the same evaluation setting. These methods include EAC \cite{li2017eac}, DSIN \cite{corneanu2018deep}, JAA \cite{shao2018deep}, LP \cite{niu2019local}, ARL \cite{shao2019facial}, SREAL \cite{li2019semantic}.

\par \textbf{Evaluation on BP4D.} Table \ref{table:BP4D} reports the F1-frame results of different methods on BP4D, the performance is evaluated for 12 action units. The results for other methods are taken from the papers. It can be seen that our method overall outperforms all previous works with an average F1-score of 63.4, compared with the previous best method, our method improves by 1.3\%.

\par \textbf{Evaluation on DISFA.} Table \ref{table:DISFA} reports the F1-frame results of different methods on BISFA, the performance is evaluated for 8 action units. The results for other methods are taken from the papers. It can be seen that our method overall outperforms all previous works with an average F1-score of 60.1, compared with the previous best method, our method improves by 1.4\%.

\subsection{Ablation Study}
To investigate the effectiveness of each component in our method, we run the same experiments using variations of the proposed network. Table \ref{table:abs} show the result of the ablation study on BP4D dataset by using the F1-score metric. Specifically, we test the contributions of the important components of our method, namely, pre-trained feature extractor (PT), region learning (RL), and transformer encoder (E). All components have two results with or without the meta-learning (ML) framework. Note that relational learning does not use the meta-learning framework for training, therefore, the difference between PT-RL-ML-E and PT-RL-E is that one imports the meta parameters of the first sub-task, while the other imports the non-meta parameters.
\par In order to clearly discuss the role of each module, we firstly look at the model results without meta-learning. Comparing the results of the baseline and pre-train model, it can be seen that the pre-train is improved by 1.3\%. Comparing the results of PT and PT-RL, it can be seen that the addition of regional learning has significantly improved by 4.2\%, indicating that AU is activated in the local region of the face. Similarly, comparing the results of PT-RL and PT-RL-E, it can be seen that the addition of relational learning has improved by 1.1\%, which further indicates that the transformer encoder can learn the AU relational information.
\par
\begin{figure}[htbp]
\centering
\includegraphics[width=0.45\textwidth]{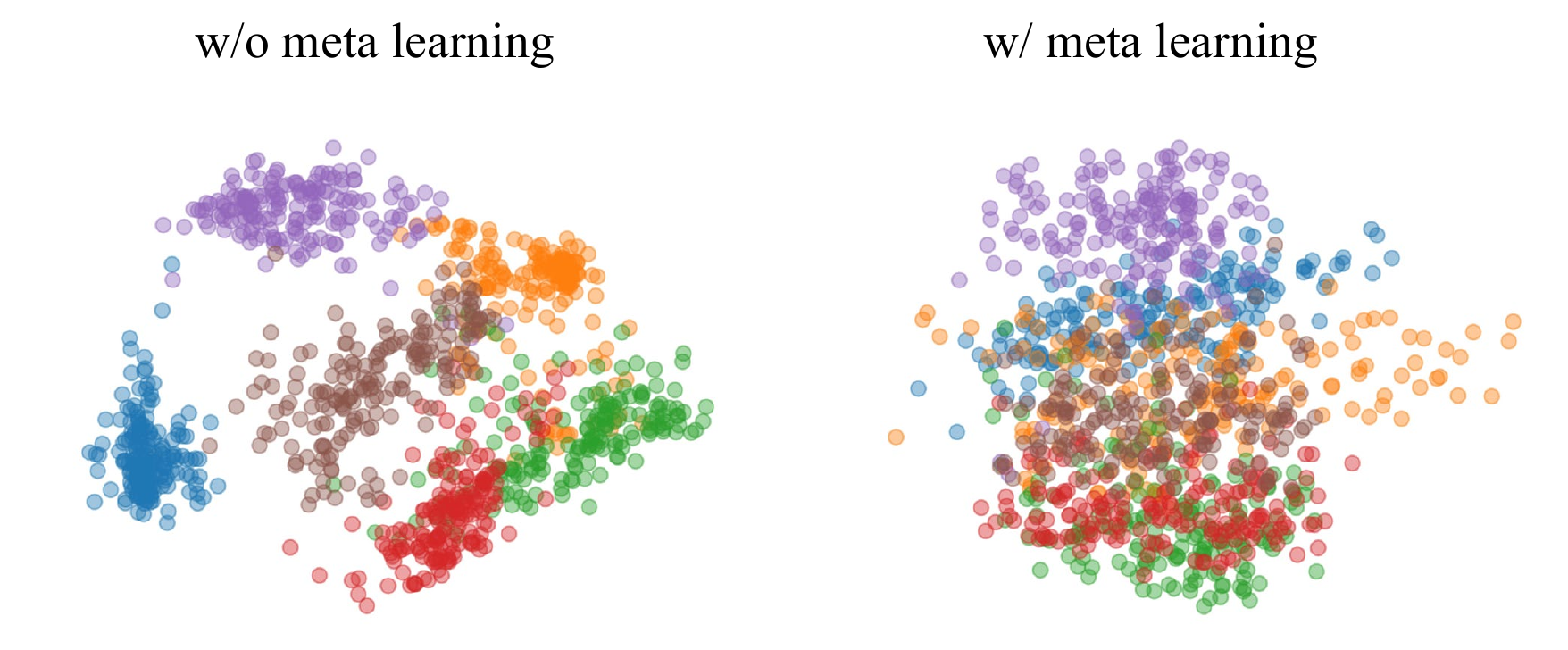}
\caption{Illustration of feature distribution of subjects without and with meta-learning. }
\label{fig:sbj}
\end{figure}

 \textbf{The impact of meta-learning on identity differences.} Comparing PT and PT-ML, PT-RL, and PT-RL-ML, the latter increased by 1.4\% and 1.0\% respectively. The experimental results show that training based on meta-learning is superior to the training based on data-driving. 
From the perspective of AU detection, it can be explained that meta-learning has learned shared information among multiple subjects, eliminating identity-caused differences. The t-SNE \cite{van2008visualizing} visualization results in Figure \ref{fig:sbj} can be clearly explained. We use t-SNE to visualize the feature distribution of 6 subjects (200 images per subject). The illustration shows that the points on the right side (with meta-learning) are more concentrated and the different points are mixed together compared to the left side (without meta-learning). This explains that meta-learning can better integrate the shared information of multiple subjects, and eliminate the identity differences. 

\section{Conclusion}
To solve the problem of the effects of identity differences on AU detection, this paper proposes a cascaded cross-subject AU detection method with meta-learning and transformer-based relation modeling. The first sub-task is AU local region representation learning based on meta-learning, which can integrate the shared information of multiple subjects and eliminate identity-caused differences. The second sub-task is relation learning based on the transformer encoder architecture, AU embeddings obtained from the first sub-task are used as input to capture AU relations. Extensive experiments prove that our model can solve the cross-subject problem well, and after adding the relation learning based on the transformer encoder architecture, it achieves the state-of-the-art results.

\section*{Acknowledgment}
This work is supported by the National Natural Science Foundation of China (No. 61503277) and the Municipal Natural Science Foundation of Tianjin (No. 20JCQNJC01230).

\bibliographystyle{plain}
\bibliography{MANet}

\end{document}